\pdfoutput=1

\documentclass[11pt]{article}

\usepackage{emnlp2021}

\usepackage{times}
\usepackage{latexsym}

\usepackage[T1]{fontenc}

\usepackage[utf8]{inputenc}

\usepackage{microtype}

\title{CodeQA: A Question Answering Dataset for Source Code Comprehension}

\author{Chenxiao Liu, Xiaojun Wan \\
  Wangxuan Institute of Computer Technology, Peking University \\
  The MOE Key Laboratory of Computational Linguistics, Peking University \\
  \texttt{\{jslcx,wanxiaojun\}@pku.edu.cn} \\
  }

\usepackage{multirow}
\usepackage{algorithm}
\usepackage[noend]{algpseudocode}
\usepackage{tabularx,booktabs}
\usepackage{listings}
\usepackage{placeins}
\usepackage{natbib}
\begin{document}

\maketitle
\begin{abstract}
We propose CodeQA, a free-form question answering dataset for the purpose of source code comprehension: given a code snippet and a question, a textual answer is required to be generated. CodeQA contains a Java dataset with 119,778 question-answer pairs and a Python dataset with 70,085 question-answer pairs. To obtain natural and faithful questions and answers, we implement syntactic rules and semantic analysis to transform code comments into question-answer pairs. We present the construction process and conduct systematic analysis of our dataset. Experiment results achieved by several neural baselines on our dataset are shown and discussed. While research on question-answering and machine reading comprehension develops rapidly, few prior work has drawn attention to code question answering. This new dataset can serve as a useful research benchmark for source code comprehension.
\end{abstract}

\section{Introduction}

Question Answering (QA) is the task of answering questions given a context about which the questions are being asked. With the advancement of deep learning and the availability of large-scale data, QA has received increasing attention from researchers. In recent years, QA has been applied into broad application domains, such as news \cite{hermann2015teaching, trischler2016newsqa}, science \cite{khot2018scitail, hardalov2020exams}, movies \cite{miller2016key}, medical field \cite{pampari2018emrqa}, etc. Among QA’s wide applications, code QA is an appealing application scenario on account of the distinctive nature of code differing from text.

In this study, we focus on generating QA pairs for source code for the purpose of source code comprehension. QA-based source code comprehension is the ability to read a code snippet and then answer questions about it, which requires understanding both source code and natural language. Take the question ``\textit{What does the code insert at the specified index?}'' together with a code snippet in Table~\ref{first-example} as an example. To answer the question, one probably first reads the source code carefully, figures out that the method adds the given parameter ``child'' at the specified index into the object ``children''. Thus one gives a proper answer: ``\textit{The given child.}''. 

\begin{table}
\small
\centering
\begin{tabular}{l}
\begin{lstlisting}[language = Java,keywordstyle = \color{blue},basicstyle = \small \ttfamily]
public void insertChildAt(Element child, int index){ 
    setChildParent(child); 
    children.add(index, child); 
}
\end{lstlisting}
\\

\textbf{Question}: What does the code insert at the specified index? \\
\textbf{Answer}: The given child. \\
\end{tabular}
\caption{\label{first-example}
A question-answer pair for a sample code snippet in the QA-based source code comprehension task. 
}
\end{table}

Compared with code summarization task \cite{haiduc2010supporting} that generates comments for codes, QA-based code comprehension task introduces more specific guidance and more explicit signals for models on what to generate. It provides more granularity levels ranging from method to variable, not just regarding several lines of code as a whole. Besides, it is easier to be evaluated since the output is more succinct, constrained and targeted \cite{kryscinski2019neural}.

QA-based source code comprehension has direct use in education to facilitate programming learning, where a system automatically answers questions about codes that someone has read. A more general use is to help improve software maintenance since it can advance the readability of code. Moreover, it can provide diverse information that can be leveraged to help perform a wide range of software engineering tasks, such as bug detection, specification inference, testing and code synthesis.

However, constructing a code QA dataset for source code comprehension is very challenging. Naturally occurring QA pairs on the Web are often complicated, noisy and contain information that cannot be inferred from the source code. We tried to collect naturally occurring QA pairs from source code management platforms like Github, QA sites for programmers like StackOverflow,  programming online judge platforms like Leetcode. But these QA pairs usually rely on knowledge apart from source code, which makes it problematic to disentangle modeling weakness from data noise. An alternative is to have experienced developers write QA pairs for source code from scratch, which is inefficient and cost-intensive. 

In this study, we introduce a new data construction process to address the above challenges and propose CodeQA, a free-form question-answering dataset. First, to ensure that QA pairs are natural and clean, we utilize code comments as data source. We pick out two large-scale well-studied datasets from Github - a Java dataset and a Python dataset. Then we select comments that are suitable to generate QA pairs from these datasets. Targeting at generating various types of QA pairs such as Wh-questions and Yes/No questions, we implement syntactic rules and semantic analysis to transform comments into QA pairs. More specifically, comments are transformed into dependency trees and converted to fit question templates that are invoked by semantic role labels. We also analyze the verbal group of comments and generate Yes/No questions. After that, QA pairs with ambiguous answers and unbalanced counts of Yes/No are filtered. 

Due to the varied nature of code comments, CodeQA covers a variety of information containing in codes, ranging from method to variable. We analyze our dataset and classify all the generated QA pairs into four categories: functionality, purpose, property and workflow. Our experiments with several baseline models demonstrate that neural models struggle to generate correct answers. These results suggest that our dataset could serve as a useful groundwork for QA-based source code comprehension.

Prior work \cite{bansal2021neural} built a dataset for code QA, but only a third of their questions are free-form. The main differences between our work and prior work are: first, all of our questions are free-form; second, our questions have diverse textual expressions and they are asking about information of various granularity in code, while questions in prior work are mostly fixed and on the single granularity.

Therefore, the contributions of this paper are as follows. 
\begin{itemize}

\item We propose the QA-based source code comprehension task and introduce a large-scale question-answering dataset containing 119,778 QA pairs for 56,545 Java codes, and 70,085 QA pairs for 44,830 Python codes. As far as we know, it is \textbf{the first} diverse free-form QA dataset specially built for source code comprehension. The dataset is available at \href{https://github.com/jadecxliu/CodeQA}{https://github.com/jadecxliu/CodeQA}. 

\item We present a data construction process to generate code QA pairs based on code comments and advance a taxonomy to classify code QA pairs into four categories.

\item We provide several baselines to evaluate the QA-based source code comprehension task. Experimental results demonstrate this dataset could serve as a useful benchmark for model and metric development.

\end{itemize}

\section{Related Work}

\subsection{Question Answering}
Question answering has a long history and has attracted increasing attention in recent years. Question Answering tasks are usually divided into four categories \cite{chen2018neural, qiu2019survey, liu2019neural}: cloze tests, multiple-choice, span prediction and free-form answering. A few examples of QA datasets in each category are CNN \& Daily Mail \cite{hermann2015teaching}, RACE \cite{lai2017race}, SQuAD \cite{rajpurkar2016squad}, MS MARCO \cite{nguyen2016ms}. Compared with other categories, free-form answering tasks show their superiority in the dimensions of understanding, flexibility, and application, which are the closest to practical application. However, the flexibility of the answer form brings difficulty to build datasets \cite{liu2019neural}. On these datasets, earlier work in question answering employed rule-based and machine-learning-based methods. Recent deep-learning techniques leveraged neural networks with different attention mechanisms and pre-trained text representation \cite{yamada2020luke, he2020deberta}, improving the ability of extracting contextual information and context-question interaction. 

In code QA, \citet{bansal2021neural} designed a context-based QA system for \textit{basic} questions about subroutines and evaluated the system by an RNN-based encoder-decoder network. They define the  ``\textit{basic}'' question as a question about a small detail of a method, such as ``What are the parameters to the method?''. These questions can be solved by parsing code without source code comprehension. Remaining questions are similar to code summarization, such as ``What does method do?''. Compared with existing work, we construct a more complex and attractive QA dataset 
for the purpose of code comprehension, which requires the understanding of both source code and natural language.
 
\subsection{Code Summarization}
Code summarization is the task of creating readable summaries that describe the functionality of a code snippet. Neural source code summarization approaches frame the problem as a sequence generation task \cite{iyer2016summarizing} and use encoder-decoder networks with attention mechanisms.
Some approaches utilized the structural information of code, such as Code2seq proposed by \citet{alon2018code2seq}, DeepCom proposed by \citet{hu2018deep}, ast-attendgr proposed by \citet{leclair2019neural}. 
Structural information can be also encoded into tree structure encoders such as Tree-LSTM \cite{shido2019automatic}, Tree-Transformer \cite{harer2019tree}, and Graph Neural Network \cite{leclair2020improved}. Besides, other techniques like 
reinforcement learning \cite{wan2018improving}, dual learning \cite{wei2019code}, retrieval-based techniques \cite{zhang2020retrieval}, language-agnostic representation learning \cite{zugner2021language} further enhance the code summarization models. 
Recently, neural architectures like Transformer \cite{vaswani2017attention} and large pre-trained models \cite{peters2018deep, radford2018improving, devlin2018bert, liu2019roberta} have brought improvements on code summarization task. Representative works are transformer designed for code \cite{ahmad2020transformer}, CodeBERT\cite{feng2020codebert}, which is a model pre-trained on the CodeSearchNet \cite{husain2019codesearchnet} dataset for programming and natural languages.

Among these noteworthy works, two datasets including a Java dataset \cite{hu2018summarizing} and a Python dataset  \cite{barone2017parallel} are popular when conducting experiments. We construct our QA dataset based on the two datasets.

\section{The Code QA Task}
In this work, we focus on building a dataset and setting up baselines for the following code QA task: Given a source code $c$ and a natural language question $q$ about $c$, a free-form textual answer $a$ is required to be generated. The textual answer $a$ may be a word, a phrase or a sentence, and it usually cannot be directly extracted from the source code. This task is different from traditional machine reading comprehension, as code is very different from text and we need programming knowledge to understand it. The source code and the natural language question are actually in two different languages and a QA system should have the ability to understand both the code language and the natural language. Moreover, the system needs to generate an answer faithful to the question and the corresponding code, rather than extract some tokens from the code. Therefore, the task is very challenging and it is very meaningful and urgent to construct and release a large-scale dataset for this task. 

\section{Dataset Construction}

\begin{table*}
\centering
\small
\begin{tabularx}{\textwidth}{p{2cm}p{3cm}XX}
\toprule
\textbf{Potential Answer} & \textbf{Question Template} & \textbf{Sample Comment} & \textbf{Generated QA}\\
\hline
subject (nsubj) & 
\textbf{Wh} mainAux otherAux verb obj modifiers? &  
The aliases will be associated with the index when it gets created. &
Q: What will be associated with the index when it gets created? \newline A: The aliases.
 \\
 \hline
direct object (dobj) & 
\textbf{Wh} mainAux nsubj otherAux verb modifiers? &
The code trims all occurrences of the supplied leading character from the given string. &
Q: What does the code trim from the given string? \newline A: All occurrences of the supplied leading character. \\
 \hline
open clausal complement (xcomp) & 
\textbf{Wh} mainAux nsubj verb modifiers? &  
The function tries to request new data if the dialog is open and a stream is set. &
Q: What does the function try if the dialog is open and a stream is set? \newline A: To request new data. \\
\hline
Temporal (TMP) & 
\textbf{When} mainAux nsubj otherAux verb obj modifiers? &   
The code takes a screenshot after every test. & 
Q: When does the code take a screenshot? \newline A: After every test. \\
\hline
Locative (LOC) &
\textbf{Where} mainAux nsubj otherAux verb obj modifiers? &
This method positions the stream at the first central directory record. &
Q: Where does this method position the stream? \newline A: At the first central directory record. \\
\hline
Manner (MNR) &
\textbf{How} mainAux nsubj otherAux verb obj modifiers? &
The code creates an instance of a class using the specified classloader. &
Q: How does the code create an instance of a class? \newline A: Using the specified classloader. \\
\hline
Cause (CAU) &
\textbf{Why} mainAux nsubj otherAux verb obj modifiers? &
The code always returns true since we wanna get all vars in scope. &
Q: Why does the code always return true? \newline A: Since we wanna get all vars in scope. \\
\hline
Purpose (PNC and PRP) &
\textbf{For what purpose} mainAux nsubj otherAux verb obj modifiers? &
The function adds and removes entries from the statements collection to munge wikibase rdf exports into a more queryable form. &
Q: For what purpose does the function add and remove entries from the statements collection? \newline A: To munge wikibase rdf exports into a more queryable form. \\
\bottomrule
\end{tabularx}
\caption{\label{template-example}
A few templates to describe the construction of QA pairs.
}
\end{table*}

\subsection{Data Source}
We construct our code QA dataset based on two code summarization datasets. The first one is a parallel corpus consisting of about a hundred thousand Python methods with descriptions written by their own programmers collected from Github \cite{barone2017parallel}. Each source code object contains a ``docstring'' (documentation string), which is retained at runtime as metadata. Programmers use docstrings to describe the functionality, interface or usage of code objects. Docstrings are extracted as natural language descriptions for code summarization tasks. The second is a parallel corpus of over seventy thousand Java code-comment pairs from Github \cite{hu2018summarizing}. The dataset contains the Java methods and the corresponding Javadoc comments. These comments describe the functionality of Java methods and are taken as code summaries.  

Code comments like Python docstrings and Javadocs can be viewed as the source of QA pairs. As code comments are deemed faithful to the code snippets, the QA pairs generated from the code comments are also faithful to the code snippets. Note that the code comments are only used for generating QA pairs, but not provided in the final code QA dataset. The comment taxonomy constructed by prior work \cite{zhai2020cpc} illustrates that the content of comment can be classified into five perspectives: \textit{what}, \textit{why}, \textit{how-it-is-done}, \textit{property} and \textit{how-to-use}. The diverse perspectives of content provide rich information to dig up and be transformed into QA pairs. Thus, we can generate question-answer pairs for code snippets by identifying potential answers in code comments, such as asking about the constraints or intentions of the key components of code occurring in comments.

\subsection{Comment Selection}
Not all comments are suitable for generating QA pairs. We thus define a selection process to help filter out noisy comments. Most comments lack the subject, such as ``attach votes count to each object of the queryset'', which starts with a verb and the hidden meaning is ``the code attaches votes count to each object of the queryset''. Incomplete sentences add difficulty to parsing in the following stage. So if a comment lack the subject, we add ``the code'' at the beginning of the sentence. Besides, we filter incomplete comments (still under development) or comments unrelated to the corresponding code. These comments are clued by keywords including ``TODO'', ``license'', ``ownership'', etc. \cite{pascarella2017classifying}.


\subsection{Question and Answer Formulation}

Typically, questions can be divided into several types \cite{day2005developing}: General questions, with Yes/No answers; Wh-questions, starting with what, where, when and so on; Choice questions, where there are some options inside the question. Since the questions in this work are converted from code comments, we focus on general questions and wh-questions, and leave choice questions as future work.

To obtain wh-questions and general questions from code comments, we implement rule-based and template-based methods to convert syntactic and semantic representations into QA pairs.

For wh-questions, we make use of dependency parsing (DP) and semantic role labeling (SRL). For one thing, we transform comments into dependency trees in the format of Universal Dependencies (UD) \cite{de2014universal} by using the allennlp parser \cite{gardner2018allennlp}. We extract a potential answer where a verb is headed by a few dependency nodes in the dependency tree with the help of semantic role labels (SRL) according to the Propbank 1.0 specifications \cite{palmer2005proposition}. The SRL model is provided by \citet{gardner2018allennlp}. For another, we extract the roles of each predicate occurred in the comment by SRL. According to Propbank, the roles are \textit{proto-agent}, \textit{proto-patient}, \textit{location}, \textit{direction}, \textit{manner}, \textit{extent}, \textit{cause}, etc. Roles like \textit{location}, \textit{direction} are classified as modifiers, which can formulate our answers. In the end, we use some of the predefined handwritten templates in \citet{dhole2020syn} to generate QA pairs. Table~\ref{template-example} presents a few templates and examples to describe the construction of questions and answers. The first three rows are from dependency heuristics and others are from SRL heuristics. The detailed process of wh-question formulation is provided in Appendix A.

With respect to general questions, we analyze the verbal group of the comment and generate a Yes/No question for every predicate that has a finite verb. We generate multiple Yes/No questions for each predicate if a comment contains multiple predicates. First, we select a clause for the current predicate and may rearrange the sequential position of semantic role labeling arguments. The standard declarative word order is preserved when generating a QA. When copular, modals, or cases 
if an auxiliary be/have/do is already present, we do not provide do-support. Otherwise, we add do-support and may move adjunct arguments relative to the main verb. As the negation label of the main verb in the verbal group indicates the polarity, according to \citet{flor2018semantic}, we do not transfer the negation into generated question, but flip the answer from ``yes/no'' to ``no/yes''. For example, from ``windows don’t have a mode like linux cli example'', we derive the question ``Do windows have a mode like linux cli example?'' and the answer ``No''.

Since some comments can not be successfully parsed to generate QA pairs, we construct 115,807 QA pairs from 44,867 code snippets in Python dataset, and 203,229 QA pairs from 56,583 code snippets in Java dataset.

\subsection{Postprocessing}
To generate high-quality code QA pairs, we filter the QA pairs that have ambiguous answers, such as answers only containing pronouns. Since some comments start with ``the method'' ``this function'' and we have added ``the code'' at the beginning of some comments when preprocessing, some generated QA pairs question about the subject and get answers like ``this method''. We also filter these QA pairs as they do not provide specific information about code snippets.

Besides, the original ratio of Yes questions to all questions is too high, with a heavily uneven proportion of Yes questions to No questions in our dataset. So we delete the majority of Yes questions to achieve a relative balance between Yes and No questions. Then we split each dataset into training, development and test sets in proportion with 8 : 1 : 1 after shuffling the pairs.

\begin{table}
\small
\centering
\begin{tabular}{l|c|c}
\toprule
 & \textbf{Java} & \textbf{Python}\\
\hline
Size of training set & 95,778 & 56,085\\
Num. of unique codes & 43,339 & 34,641 \\
Num. of unique tokens in codes & 27,504 & 108,571 \\
Num. of unique tokens in questions & 13,097 & 11,401 \\
Num. of unique tokens in answers & 13,820 & 12,723 \\
Avg. Num. of tokens per code & 119.52 & 48.97 \\
Avg. Num. of tokens per question & 9.48 & 8.15 \\
Avg. Num. of tokens per answer & 4.74 & 4.07 \\
\hline
Size of dev set & 12,000 & 7,000 \\
Size of test set & 12,000 & 7,000 \\
\bottomrule
\end{tabular}
\caption{\label{dataset-statistics}
CodeQA dataset statistics.
}
\end{table}

\begin{table}
\small
\centering
\begin{tabular}{l|c|c}
\toprule
 & \textbf{Java} & \textbf{Python}\\
\hline
Dev answer repetition & \multirow{2}{*}{5.51\%} & \multirow{2}{*}{1.17\%}\\
rate against Train & & \\
Test answer repetition  & \multirow{2}{*}{3.63\%} & \multirow{2}{*}{1.56\%} \\
rate against Train & & \\
Span answer & 1.47\% & 4.99\% \\
Extraction answer & 3.71\% & 12.57\% \\
\bottomrule
\end{tabular}
\caption{\label{repetition}
The repetition rate and extraction rate of answers.
}
\end{table}

\begin{table*}
\centering
\small
\begin{tabularx}{\textwidth}{p{2cm}Xp{1.6cm}<{\centering} }
\toprule
\textbf{Type} & \textbf{Example QA} & \textbf{Percentage}\\
\hline
Functionality & 
Q: What does the code instantiate? \newline A: An input data placeholder variable.
& 63\% \\
\hline
Purpose & 
Q: For what purpose does this method return a new name? \newline A: To address the key value pair in the context of that user.
& 9\%\\
\hline
Property &
Q: When is this event triggered on the slave instances? \newline A: Every time a stats report is to be sent to the locust master.
& 17\%\\
\hline
Workflow & 
Q: How does a maximal independent set of nodes in g return? \newline A: By repeatedly choosing an independent node of minimum degree.
& 11\%\\
\bottomrule
\end{tabularx}
\caption{\label{categorizations}
Distribution of different QA pairs among 100 randomly chosen samples.
}
\end{table*}

\begin{table}
\centering
\small
\begin{tabular}{l|c}
\toprule
\textbf{Question Type} & \textbf{Percentage} \\
\hline
What & 67.24\% \\
How & 8.93\%\\
Where & 5.85\%\\
When & 6.89\%\\
Why & 1.02\% \\
For what purpose & 5.08\%\\
Yes/No & 2.86\%\\
Other & 2.13\%\\
\bottomrule
\end{tabular}
\caption{\label{question-contains}
Distribution of questions after automatic partitioning.
}
\end{table}

\section{Dataset Analysis}

In this section, we introduce the overall statistics of our dataset and verify the free-form characteristic of our dataset. To explore the distribution of different kinds of source code QA pairs, we propose a taxonomy of four types of source code comprehension.

\subsection{Overall Statistics}
Table~\ref{dataset-statistics} describes the basic statistics of our CodeQA dataset. In Java dataset, there are 95,778 training pairs, 12,000 development pairs and 12,000 test pairs. In Python dataset, there are 56,085 training pairs, 7,000 development pairs and 7,000 test pairs. 

We calculate the percentage of answers in the development or test set that can be found in the training set (excluding Yes/No answers). Besides, we calculate the percentage of answers in all sets that are spans of the code, and the percentage of answers whose each token could be extracted from the code, as shown in Table~\ref{repetition}. The statistics attest to the free-form nature of our dataset.

\begin{table*}
\centering
\small
\begin{tabular}{l|ccccc|ccccc}
\toprule
\multirow{2}{*}{} & \multicolumn{5}{c|}{Dev} & \multicolumn{5}{c}{Test} \\
\cline{2-11}
& BLEU & ROUGE & METEOR & EM & F1 & BLEU & ROUGE & METEOR & EM & F1 \\
\hline
Seq2seq & 28.86 & 20.13 & 5.21 & 4.11 & 21.02 & 28.29 & 18.88 & 4.79 & 3.23 & 19.75  \\
Dual Encoder & 30.80 & 22.78 & 7.21 & 5.95 & 23.57 & 29.51 & 20.39 & 6.27 & 4.06 & 21.16  \\
Transformer & 31.54 & 23.67 & 7.69 & 6.52 & 24.39 & 30.22 & 21.26 & 6.77 & 4.41 & 22.04  \\
CodeBERT & 33.45 & 31.80 & 11.57 & 7.80 & 32.69 & 32.40 & 28.22 & 10.10 & 6.20 & 29.20  \\
\hline
\hline
CodeBERT$^{*}$ & 37.32 & 33.06 & 13.36 & 11.00 & 34.38 & 35.19 & 33.56 & 10.99 & 9.00 & 32.07  \\
Human$^{*}$ & 64.18 & 66.10 & 29.04 & 37.00 & 63.22 & 62.97 & 63.44 & 28.19 & 34.00 & 64.06 \\
\bottomrule
\end{tabular}
\caption{\label{java-automatic-evaluation}
Performance of various models on Java dataset. $^{*}$\textit{means the result is obtained on only 100 questions sampled in the respective dataset. We evaluate the 100 answers generated by CodeBERT and that given by an experienced programmer respectively. The results are used only for comparing CodeBERT and Human.}
}
\end{table*}

\begin{table*}
\centering
\small
\begin{tabular}{l|ccccc|ccccc}
\toprule
\multirow{2}{*}{} & \multicolumn{5}{c|}{Dev} & \multicolumn{5}{c}{Test} \\
\cline{2-11}
& BLEU & ROUGE & METEOR & EM & F1 & BLEU & ROUGE & METEOR & EM & F1 \\
\hline
Seq2seq & 29.70 & 22.18 & 6.36 & 2.95 & 23.67 & 30.09 & 21.69 & 6.38 & 2.84 & 23.27 \\
Dual Encoder & 28.57 & 17.62 & 4.43 & 2.22 & 18.78 & 28.90 & 17.79 & 4.53 & 2.22 & 18.97 \\
Transformer & 30.85 & 23.96 & 7.91 & 3.37 & 25.30 & 30.69 & 23.26 & 7.80 & 3.35 & 24.59 \\
CodeBERT & 33.84 & 30.27 & 11.51 & 5.50 & 31.57 & 34.86 & 30.28 & 12.51 & 4.93 & 31.56 \\
\hline
\hline
CodeBERT$^{*}$ & 34.02 & 29.89 & 12.18 & 8.00 & 35.89 & 34.78 & 30.93 & 13.45 & 8.00 & 34.10  \\
Human$^{*}$ & 56.21 & 60.65 & 26.32 & 34.00 & 61.98 & 55.96 & 59.55 & 25.49 & 32.00 & 61.59 \\
\bottomrule
\end{tabular}
\caption{\label{python-automatic-evaluation}
Performance of various models on Python dataset.
$^{*}$\textit{ has the same meaning as in the above table.}
}
\end{table*}

\subsection{Categorization}

We are not aware of an agreed-upon typology of all code QA types. Categorizations of different types of code summarization exist \cite{pascarella2017classifying, zhai2020cpc}, but the provided categorizations differ and are manually classified by coders in general. \citet{bansal2021neural} generated QA pairs about code and divided the questions into six types, including basic extractive question types like ``What is the return type of method?'', ``What are the parameters of method?'' and question types equivalent to code summarization, i.e. ``What does method do?''. After consulting both prior works and a separate part of the training data, we characterize the data into the following four types. 

These types consist of (1) Functionality. It provides a definition of the range of functions that the subject and/or its interface can perform. (2) Purpose. It explains the reason why the subject is provided or the design rationale of the subject. (3) Property. It declares properties of the subject, such as pre-conditions and post-conditions of a method or some statements. Pre-conditions specify the constraints that should satisfy in order to use the subject while post-conditions indicate the result of using the subject. (4) Workflow. It describes how the subject is done, which means implementation details like the design or the workflow of the subject. The subject mentioned in the four types can either be the whole method or a key component of the code, e.g. a statement, a variable.

To get a better understanding of the categorizations of code QA pairs, we sampled 100 QA pairs in the development set of Python, and then manually labeled the examples with the categories shown in Table~\ref{categorizations}. The results show that more than half of questions target functionality, while 37\% questions ask about purpose, property, or workflow.

To show the diversity in QA pairs, we also automatically categorize all the QA pairs in 
Table~\ref{question-contains}. We can see that What question makes up 67.24\% of the data; 27.77\% of the questions are How/Where/When/Why/For what purpose; 2.86\% are Yes/No; and the remaining 2.13\% are other types.

\section{Experiments}

In this section, we first introduce four baseline models for this task. Then we compare and analyze the results of different models under both automatic and human evaluation metrics. 

\subsection{Baselines}

We present baseline results on CodeQA by examining four existing typical approaches. Since no previous work specifically designs a model for QA-based source code comprehension, we make some modifications to each of the existing approaches. Note that we do not employ retrieval models, for the answer repetition rate is quite low as shown in Table~\ref{repetition}. Details about hyperparameter settings of all baselines are provided in the Appendix B. 

\begin{itemize}
\item \textbf{Seq2seq}: A Seq2seq model \cite{sutskever2014sequence} with attention and copy mechanism \cite{see2017get}. While originally designed for text-to-text generation, it is commonly used in free-form question-answering as well \cite{nguyen2016ms}. The input of the model is in the form of ``[CLS] Question [SEP] Code''. Since models using original code tokens could perform better than models using abstract syntax tree (AST) sequences \cite{ahmad2020transformer}, we employ the code tokens as input for all baseline models.

\item \textbf{Dual Encoder}: A seq2seq model with two encoders. The model first builds a code representation and a question representation by its code-info encoder and question-info encoder respectively. After that, it concatenates the two representations for the decoder. Both encoders and decoder are similar to the architecture in the above Seq2seq model.

\item \textbf{Transformer}: A Transformer encoder-decoder model \cite{vaswani2017attention} with relative position representations \cite{shaw2018self} and copy attention. The input is a sequence containing a question and a code separated by [SEP]. Since the semantic representation of a code does not rely on the absolute positions of its tokens, the Transformer ignores the directional information and encodes pairwise relationship \cite{ahmad2020transformer}.  

\item \textbf{CodeBERT}: A Transformer encoder-decoder model where the encoder are initialized with CodeBERT \cite{feng2020codebert}. Following BERT \cite{devlin2018bert} and RoBERTa \cite{liu2019roberta}, CodeBERT is a bimodal model pre-trained with natural language and programming languages including Python and Java, etc. We fine-tune the model parameters on our dataset and predict answers given an input sequence consisting of a question and a code. Note that when training a version of CodeBERT by traversing the AST of code, model does not bring improvements on generation tasks \cite{feng2020codebert}. Thus we do not transform the code into tree structure.
\end{itemize}

\begin{table*}
\centering
\small
\begin{tabular}{l|cc|cc|cc|cc}
\toprule
\multirow{3}{*}{} & \multicolumn{4}{c|}{\textbf{Java}} & \multicolumn{4}{c}{\textbf{Python}} \\
\cline{2-9}
& \multicolumn{2}{c|}{Fluency} & \multicolumn{2}{c|}{Correctness} & \multicolumn{2}{c|}{Fluency} & \multicolumn{2}{c}{Correctness}\\
\cline{2-9}
& Dev & Test & Dev & Test & Dev & Test & Dev & Test\\
\hline
Seq2seq & 2.09 & 2.21 & 1.66 & 1.62 & 2.23 & 2.24 & 1.61 & 1.63  \\
Dual Encoder & 2.23 & 2.19 & 1.79 & 1.57 & 2.38 & 2.44 & 1.58 & 1.64  \\
Transformer & 2.37 & \textbf{2.41} & 1.84 & 1.75 & 2.40 & 2.44 & 1.70 & 1.72  \\
CodeBERT & \textbf{2.53} & 2.40 & \textbf{1.92} & \textbf{1.89} & \textbf{2.45} & \textbf{2.50} & \textbf{1.78} & \textbf{1.79}  \\
\bottomrule
\end{tabular}
\caption{\label{human-evaluation}
Human evaluation results. Scores of each aspect range from 1 to 3 and higher scores are better.
}
\end{table*}

\subsection{Automatic Evaluation Metrics}
The model output is evaluated using several automatic metrics: BLEU \cite{papineni2002bleu}, ROUGE-L \cite{lin2004rouge}, METEOR \cite{banerjee2005meteor}, Exact Match (EM) and F1. 

\subsection{Model Performance}
Tables~\ref{java-automatic-evaluation},~\ref{python-automatic-evaluation} present the results with different models for the QA-based source code comprehension task on Java and Python datasets, respectively. CodeBERT performs the best, followed by Transformer. It is not surprising since pre-trained models on programming codes and texts are more powerful in encoding code representations and bridging the gap between code language and natural language, thus improving the performance. We see that all the models get low Exact Match scores, indicating that answering the questions with the same token string as the gold answer is rather difficult.

\begin{table}
\centering
\small
\begin{tabular}{l}
\begin{lstlisting}[language = Java,keywordstyle = \color{blue},basicstyle = \small \ttfamily]
public IOContainer append(IOObject[] output){ 
    List<IOObject> newObjects = new LinkedList<>();
    for(int i = _NUM; i < output.length; i++){ 
        newObjects.add(output[i]); 
    } 
    newObjects.addAll(ioObjects); 
    return new IOContainer(newObjects); 
}
\end{lstlisting}
\\
\textbf{Q}: What is added to the given ioobjects? \\
\textbf{A}: all ioobjects of this container \\
Seq2seq: an array \\
Dual Encoder: the given objects \\
Transformer: a new array \\
CodeBERT: an iocontent container \\
\begin{lstlisting}[language = Python,keywordstyle = \color{blue},basicstyle = \small \ttfamily]
def get_page_args():
    pages = {}
    for arg in request.args:
        re_match = re.findall("page_(.*)", arg)
        if re_match:
            pages[re_match[0]] = int(request.args.get(arg))
    return pages
\end{lstlisting}
\\
\textbf{Q}: What does the code get? \\
\textbf{A}: page arguments\\
Seq2seq: the page \\
Dual Encoder: the args \\
Transformer: the page of the pages \\
CodeBERT: page arguments from the request \\
\end{tabular}
\caption{\label{running-example}
Examples of different models’ performance on Java and Python datasets.
}
\end{table}

\subsection{Human Performance}
We assess human performance on CodeQA’s development and test sets. Due to the large scale of the dataset, we sampled 100 questions from each set and asked two experienced programmers to give answers according to code snippets on two sets respectively. To make a comparison between model performance and human performance, we pick up the best model (CodeBERT) and evaluate its output on the same 100 questions. As shown in the last two rows of Tables~\ref{java-automatic-evaluation},~\ref{python-automatic-evaluation}, the model’ performance has a significant gap compared with human's.


\subsection{Human Evaluation}
Besides automatic evaluation, we randomly sampled 100 QA pairs from the development and test sets of CodeQA respectively, and asked two programmers to evaluate outputs of baselines on two sets respectively in the following aspects. \textit{Fluency} measures if an answer is grammatically correct and is fluent to read. \textit{Correctness} measures if an answer is targeting the given question and code. Each reviewer gives a score between 1 and 3 for each aspect, with 3 indicating the best quality. 

As shown in Table~\ref{human-evaluation}, CodeBERT gets the competitive performance in most metrics. All models get relatively poor performances on the aspect of correctness compared with fluency. The low scores of correctness indicate that it is quite challenging for models to do well in code QA.

\subsection{Qualitative Analysis}

We provide a couple of examples in Table~\ref{running-example} to demonstrate the outputs from different baselines (more qualitative examples are provided in Appendix C). In the Java example, CodeBERT captures the key component ``io container'' while other models generate imprecise concepts. As the Python code tries to get all arguments of a page, the first three baselines generate answer either about ``page'' or about ``args'' while CodeBERT contains both concepts. The examples reveal that, in comparison to the Seq2seq model and the Transformer model, CodeBERT generates more detailed and accurate answers.

\section{Conclusion}

In this paper, we 
build the first diverse free-form question answering dataset for code by transforming code comments into QA pairs. We also provide several neural baselines, and demonstrate that CodeQA could lay the foundation for further research on QA-based source code comprehension.

For future work, how to expand the number of question types and generate more high-quality QA pairs are the major challenges. Besides, we will explore more powerful QA models to better leverage information from code and capture interaction between code and question.

\section*{Acknowledgments}

This work was supported by National Natural Science Foundation of China (61772036), Beijing
Academy of Artificial Intelligence (BAAI) and Key
Laboratory of Science, Technology and Standard in
Press Industry (Key Laboratory of Intelligent Press
Media Technology). We appreciate the anonymous
reviewers for their helpful comments. Xiaojun Wan
is the corresponding author.

\bibliography{ref}
\bibliographystyle{acl_natbib}

\clearpage
\appendix


\section{Wh-question Formulation Details}
\label{sec:appendix}

To generate wh-questions, we first parse comments into dependency trees in the format of Universal Dependencies (UD) by using the allennlp parser. Dependency trees are syntactic tree structures, where syntactic units are connected via links. We extract the clause of a verb headed by a few dependency nodes which can serve as answers with the help of PropBank's predicate-argument structure (SRL). The clause is treated as a combination of a subject, an object, the head verb and other non-core arguments. Furthermore, the clause can be refined with modals, auxiliaries and negations if found around the verb. The SRL model is provided by Gardner et al. (2018). Then templates in Dhole and Manning (2020) are used to generate QA pairs. The templates convert \textit{What} to \textit{Who/Whom}, \textit{When} or \textit{Where} depending on the named entity of the answer. To ensure subject-predicate concord, templates modify \textit{do} to \textit{does} or \textit{did} relying on the tense and number of the subject. 

Algorithm 1 illustrates the heuristic rules of dependency parsing.

\begin{algorithm}[h]
\caption{Heuristic Rules of DP} 
\begin{algorithmic}[1]
\State $ \{d_{0},...,d_{n}\} \leftarrow DP(w_{0}...w_{n}) $
\For{$ i = 0\ to\ n $} 
\If{$parent(d_{i}) \ is \ not \ null$} 
\State $d_{v} \leftarrow parent(d_{i})$
\State $\{A_{0},...,A_{CAU},A_{TMP}\} \leftarrow SRL(d_{v})$
\State $subj \leftarrow A_{0}$
\If{$d_{i} \in A_{1}$} 
\State $obj \leftarrow A_{1}$
\Else
\State $obj \leftarrow A_{2}$
\EndIf
\State $ A_{x} \leftarrow \sum(A_{3},...,A_{TMP})$
\State $verb \leftarrow \{d_{v}, modals, negation\}$
\State $template \leftarrow dep_{type} \leftarrow d_{i}$
\State $QA \leftarrow template(subj, obj,$
\Statex \qquad \qquad \qquad \qquad $ A_{x}, verb)$
\EndIf
\EndFor

\end{algorithmic}
\end{algorithm}

Then we extract the roles of each predicate occurred in the comment by the SRL model provided by Gardner et al. (2018). Semantic roles include the generalized core arguments of predicates labeled as A0, A1, etc., with a set of adjunct modifiers. According to Propbank 1.0, the roles are \textit{proto-agent}, \textit{proto-patient}, \textit{location}, \textit{direction}, \textit{manner}, \textit{extent}, \textit{cause}, etc. Roles like \textit{location}, \textit{direction} are classified as modifiers, which can formulate our answers. We make use of a set of predefined handwritten templates in Dhole and Manning (2020), which convert a comment into an interrogative statement by rearranging the arguments according to the modifier. 

Algorithm 2 describes the heuristic rules of semantic role labeling.

\begin{algorithm}[h]
\caption{Heuristic Rules of SRL} 
\begin{algorithmic}[1]
\State $ \{SRL_{0},...,SRL_{s}\} \leftarrow SRL(w_{0}...w_{n}) $
\For{$ i = 0\ to\ s $} 
\If{$SRL_{i} \ contains \ A_{0} \ or A_{1} \ and  \geq 1 \ A_{m}$} 
\State $\{A_{0},...,A_{CAU},A_{TMP}\} \leftarrow SRL_{i}$
\If{$A_{x} = modifier$} 
\For{$ A_{x} \in SRL_{i}$} 
\State $subj \leftarrow A_{0}$
\State $ A_{x}^{-} \leftarrow \sum(A_{3},...,$
\Statex \qquad \qquad \qquad \qquad $A_{TMP} - A_{x})$
\State $verb \leftarrow \{A_{v}, modals, $
\Statex \qquad \qquad \qquad \qquad $negation\}$
\State $template \leftarrow modifier_{type} $
\Statex \qquad \qquad \qquad \qquad $\leftarrow A_{x}$
\State $QA \leftarrow template(subj, A_{x},$
\Statex \qquad \qquad \qquad \qquad $ A_{x}^{-}, verb)$
\EndFor
\EndIf
\EndIf
\EndFor

\end{algorithmic}
\end{algorithm}

\section{Baseline Details}

\begin{itemize}

\item \textbf{Seq2seq}: bidirectional RNN with number of layers = 2, hidden size = 512, batch size = 32, beam size = 4, learning rate = 0.002, dropout = 0.2, Adam optimizer (Kingma and Ba, 2014). 

\item \textbf{Dual Encoder}: Both the code-info encoder and the question-info encoder have 2 layers. Other hyper-parameters are same as Seq2seq.

\item \textbf{Transformer}: Transformer model with number of layers = 6, number of heads = 8, hidden size = 512, batch size = 32, beam size = 4, initial learning rate = 0.0001, dropout = 0.2, Adam optimizer.

\item \textbf{CodeBERT}: The encoder is the pre-trained CodeBERT, while the decoder is a transformer structure with number of layers = 6, number of heads = 12. Other hyper-parameters: batch size = 64, beam size = 10, learning rate = 5e-5, Adam optimizer.

\end{itemize}

For each of the first three models, we train the model for a maximum of 200 epochs on a Nvidia 1080 Ti GPU and perform early stop if the validation performance does not improve for 20 consecutive iterations. We fine-tune CodeBERT for 20 epochs on 3 Nvidia 1080 Ti GPUs and select the checkpoint with best BLEU score.

\section{Qualitative Examples}

\begin{table*}
\centering
\small
\begin{tabular}{l}
\hline 
\begin{lstlisting}[language = Java,keywordstyle = \color{blue},basicstyle = \small \ttfamily]
public Select<T> sortAsc(String[] columns){ 
    for(String column:columns){ 
        mSortingOrderList.add(column + STRING); 
    } 
    return this; 
}
\end{lstlisting}
\\
\textbf{Question}: How does the specified columns sort? \\ 
\textbf{Answer}: in asc order \\
Seq2seq: in ascending order \\
Dual Encoder: in desc order \\
Transformer: in desc order \\
CodeBERT: in ascending order \\
\hline 
\begin{lstlisting}[language = Java,keywordstyle = \color{blue},basicstyle = \small \ttfamily]
public static int count(String string, String mark){ 
    if(!TextUtils.isEmpty(string) && !TextUtils.isEmpty(mark)){
        int count = _NUM; 
        int index = string.indexOf(mark);
        while(index != -_NUM){ 
            count++; 
            string = string.substring(index + mark.length()); 
            index = string.indexOf(mark); 
        } 
        return count; 
    } 
    return _NUM; 
}
\end{lstlisting}
\\
\textbf{Question}: What does the code count?  \\
\textbf{Answer}: how many marks existed in string \\
Seq2seq: the string \\
Dual Encoder: the number of elements in the string \\
Transformer: the number of occurrences of this string \\
CodeBERT: the number of times the given string \\
\hline 
\begin{lstlisting}[language = Java,keywordstyle = \color{blue},basicstyle = \small \ttfamily]
public synchronized void create(long seqno)
        throws ReplicatorException, InterruptedException{ 
    if(file.exists()){ 
        throw new THLException(STRING + file.getName());
    } 
    try{ 
        dataOutput = new BufferedFileDataOutput(file, bufferSize); 
    } catch(IOException e){ 
        throw new THLException(STRING + file.getName(), e); 
    } 
    mode = AccessMode.write; 
    try{
        write(MAGIC_NUMBER); 
        write(MAJOR_VERSION); 
        write(MINOR_VERSION); 
        write(seqno);
        flush(); 
    } catch(IOException e){ 
        throw new THLException(STRING + file.getName(), e); 
    } 
    baseSeqno = seqno; 
    if(logFlushTask != null) logFlushTask.addLogFile(this); 
}
\end{lstlisting}
\\
\textbf{Question}: What does the code create?  \\
\textbf{Answer}: a new log file \\
Seq2seq: a new instance \\
Dual Encoder: a flie \\
Transformer: a file \\
CodeBERT: a log file \\
\hline 
\end{tabular}
\caption{\label{java-example}
Qualitative examples of different models’ performance on Java dataset.
}
\end{table*}

\begin{table*}
\centering
\small
\begin{tabular}{l}
\hline 
\begin{lstlisting}[language = Python,keywordstyle = \color{blue},basicstyle = \small \ttfamily]
def html_to_xhtml(html):
    try:
        html = html.getroot()
    except AttributeError:
        pass
    prefix = "{%s}" % XHTML_NAMESPACE
    for el in html.iter(etree.Element):
        tag = el.tag
        if tag[0] != '{':
            el.tag = prefix + tag
\end{lstlisting}
\\
\textbf{Question}: How do all tags in an html tree convert to xhtml?  \\
\textbf{Answer}: by moving them to the xhtml namespace \\
Seq2seq: recursively \\
Dual Encoder: with the given html \\
Transformer: using xhtml tags \\
CodeBERT: by removing their xhtml namespace \\
\hline 
\begin{lstlisting}[language = Python,keywordstyle = \color{blue},basicstyle = \small \ttfamily]
def table_extend(tables, keep_headers=True):
    from copy import deepcopy
    for ii, t in enumerate(tables[:]): 
        t = deepcopy(t)
        if t[0].datatype == 'header':
            t[0][0].data = t.title
            t[0][0]._datatype = None
            t[0][0].row = t[0][1].row
            if not keep_headers and (ii > 0):
                for c in t[0][1:]:
                    c.data = ''
        if ii == 0:
            table_all = t
        else:
            r1 = table_all[-1]
            r1.add_format('txt', row_dec_below='-')
            table_all.extend(t)
    table_all.title = None
    return table_all
\end{lstlisting}
\\
\textbf{Question}: What does the code extend? \\ 
\textbf{Answer}: a list of simple tables \\
Seq2seq: a table \\
Dual Encoder: a list of 0 \\
Transformer: the tables \\
CodeBERT: a list of tables \\
\hline 
\begin{lstlisting}[language = Python,keywordstyle = \color{blue},basicstyle = \small \ttfamily]
def calc_angle(v1, v2, v3):
    v1 = v1 - v2
    v3 = v3 - v2
    return v1.angle(v3)
\end{lstlisting}
\\
\textbf{Question}: What does the code calculate?  \\
\textbf{Answer}: the angle between 3 vectors representing 3 connected points \\
Seq2seq: the angle of the v1 \\
Dual Encoder: the v2 \\
Transformer: the angle angle of an error angle \\
CodeBERT: the angle between two numbers \\
\hline 

\end{tabular}
\caption{\label{python-example}
Qualitative examples of different models’ performance on Python dataset.
}
\end{table*}

\end{document}